\documentclass[sigconf]{acmart}

\usepackage{amsmath,amsfonts}
\usepackage{algorithmic}
\usepackage{graphicx}
\usepackage{xcolor}
\begin{document}

\title{RouteCost: A Production-Inspired Multi-Stage Framework for Pre-Order Shipping Cost Estimation in E-Commerce}

\author{Xianling Zeng}
\authornote{Corresponding author.}
\affiliation{%
  \institution{College of Engineering, Northeastern University}
  \city{Boston}
  \country{United States}
}
\email{zeng.xian@northeastern.edu}

\author{Zihan Yu}
\affiliation{%
  \institution{College of Professional Studies, Northeastern University}
  \city{Boston}
  \country{United States}
}
\email{yu.zihan1@northeastern.edu}

\author{Sichen Zhao}
\affiliation{%
  \institution{College of Engineering, Northeastern University}
  \city{Boston}
  \country{United States}
}
\email{zhao.siche@northeastern.edu}

\author{Yalun Qi}
\affiliation{%
  \institution{Khoury College of Computer Science, Northeastern University}
  \city{Boston}
  \country{United States}
}
\email{qi.yal@northeastern.edu}

\author{Zhiming Xue}
\affiliation{%
  \institution{College of Engineering, Northeastern University}
  \city{Boston}
  \country{United States}
}
\email{xue.zh@northeastern.edu}

\begin{abstract}
Accurate pre-order shipping cost estimation is important in e-commerce because it affects price presentation, margin planning, and conversion. In practice, shipping cost is shaped not only by distance but also by destination demand mix, billable weight, dimensional pricing, surcharge triggers, and latent operational effects such as shipment consolidation. Static lookup methods therefore miss important sources of variation, while monolithic regressors may exploit strong but non-causal correlations. We propose RouteCost, a production-inspired multi-stage framework that decomposes the problem into time-aware demand forecasting, fee-card-informed baseline pricing, Stage 2 residual correction, and proxy-based box-consolidation inference. Route-level cost estimates are aggregated through a route-weighted expectation formulation to produce product-level shipping cost predictions. Across over 250,000 orders, 260 products, and 18 months of order history, the framework improves predictive quality and aggregate calibration while preserving route-level interpretability.
\end{abstract}



\maketitle
\section{Introduction}
Pre-order shipping cost estimation is a foundational capability in modern e-commerce systems. Before an order is placed, retailers often need fulfillment-cost estimates to support price display, margin control, promotional planning, and customer-facing delivery commitments. In small-parcel environments, this task is shaped not only by geographic distance, but also by carrier pricing rules, dimensional billing, destination-specific demand patterns, and operational effects that may not be directly observable at prediction time \cite{ref1,ref2}. In practice, these factors are also subject to real-world distribution shift. Destination demand mix can change over time because of seasonality---for example, outdoor products may have weak demand in northern regions during winter but experience a sharp increase in spring---while carrier pricing conditions may also shift due to holiday peak surcharges, fee updates, or rate-card revisions. These changes make pre-order shipping cost estimation not only a prediction problem, but also a robustness challenge in a changing operational environment.

A natural first approach is a static lookup based on carrier rate tables and simple shipment attributes. Such methods are operationally useful but too rigid to capture variation introduced by destination mix, surcharge behavior, and hidden consolidation effects. A second approach is to fit a single regression model over a large feature set. However, in preliminary modeling attempts, some business variables exhibited strong empirical correlation with realized shipping cost without constituting causal pricing drivers. For example, larger products often also have higher wholesale cost, so wholesale cost and parcel cost can appear positively correlated across categories such as stools, chairs, sofas, and beds. A monolithic model may therefore learn that higher wholesale cost implies higher shipping cost, even though carrier charges are driven by physical shipment characteristics such as size, weight, and surcharge triggers rather than merchandise value itself. Two rugs with identical shipment dimensions but very different wholesale costs---for instance, a low-cost rug and a hand-knotted Persian rug---should incur essentially the same shipping cost. Using wholesale cost as a predictive driver would therefore introduce misleading correlations, improving in-sample fit while weakening interpretability and robustness under distribution shift \cite{ref3,ref4}. More broadly, an end-to-end model may entangle stable pricing logic with shifting demand patterns and latent operational effects. As a result, changes such as seasonal demand reallocation or surcharge updates can trigger broader retraining even when the core pricing mechanism remains unchanged.

Rather than treating shipping cost prediction as a single black-box regression problem, RouteCost formulates it as a route-weighted expectation problem. Under our simplified fulfillment setting - single origin warehouse, single parcel carrier, and standard service level - each product is associated with a finite set of destination-zone routes, and expected shipping cost is computed as a weighted sum of route-level cost estimates. This structured design avoids forcing a single model to absorb all sources of variation at once. Instead, it separates demand estimation, pricing structure, residual correction, and latent consolidation effects into distinct components, which can improve interpretability and support more targeted updates when operational conditions change.

We make three contributions: (1) we formulate pre-order shipping cost estimation as a route-weighted expectation problem; (2) we introduce a multi-stage architecture that separates demand forecasting, baseline pricing, residual correction, and proxy-based box-consolidation inference; and (3) we show through temporal backtesting and route-level decomposition that structured decomposition improves both predictive quality and interpretability.

\section{Related Work}
Research related to this work lies at the intersection of logistics pricing, demand forecasting, and structured machine learning for operational decision support. Learning-based methods are increasingly used to complement classical routing and transportation planning frameworks, especially in settings where demand and operational conditions evolve over time \cite{ref1,ref5}. While our work does not directly solve a vehicle routing problem, it shares the same motivation of using data-driven estimation to support downstream logistics decisions.

Prior work has also applied supervised learning to transportation and fulfillment-related tasks, often finding that hybrid designs combining domain structure with learned corrections outperform purely rule-based systems in noisy operational environments \cite{ref2,ref6}. This motivates our decision to preserve explicit pricing structure in Stage 1 while using Stage 2 to model residual nonlinear error. Consolidation has long been recognized as an important logistics strategy because combining multiple shipments can reduce transportation cost and exploit scale economies \cite{ref7,ref8}. In many pre-order settings, however, consolidation outcomes are not directly observable at prediction time, which motivates our proxy-based treatment.

\section{Method}
\subsection{Framework Overview}
\begin{figure}[!t]
\centering
\includegraphics[width=\columnwidth]{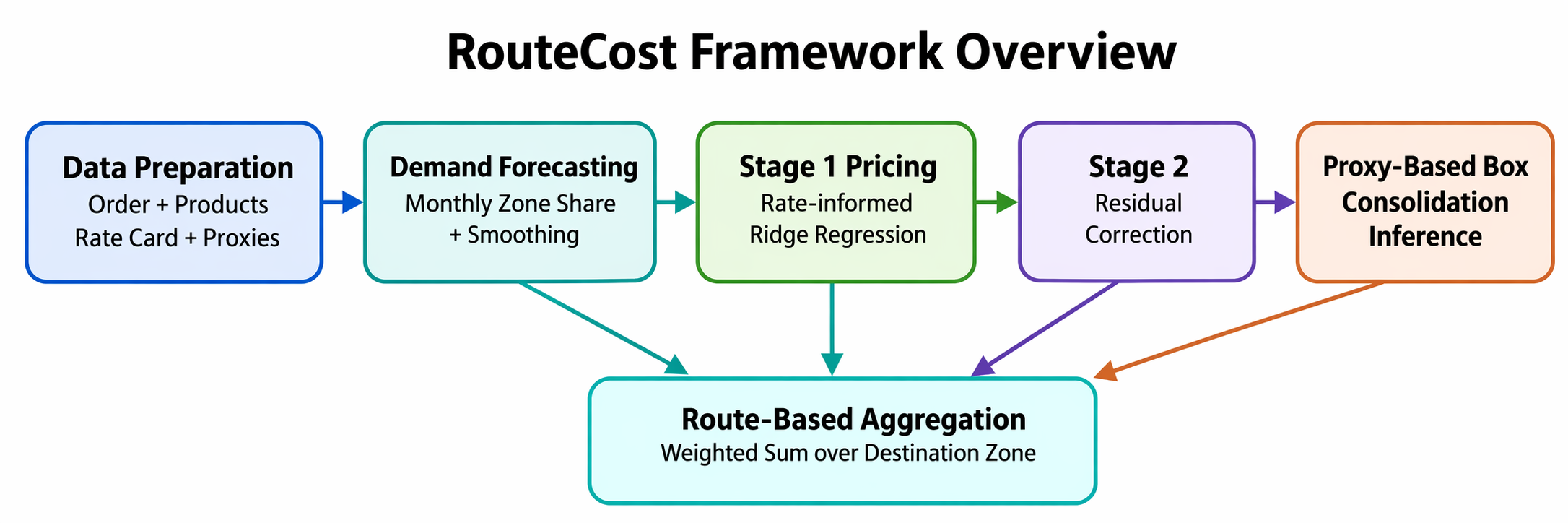}
\caption{Route Cost framework overview.}
\label{fig:framework}
\end{figure}

As shown in Figure~\ref{fig:framework}, the RouteCost framework contains four modules: (1) time-aware demand forecasting, which estimates route weights by forecasting destination-zone shares over time; (2) fee-card-informed Stage 1 pricing, which produces a structured baseline estimate using billable weight, dimensions, rate-card lookups, and surcharge features; (3) Stage 2 residual correction, which captures nonlinear error not explained by the baseline; and (4) proxy-based box-consolidation inference, which estimates latent savings associated with likely consolidation opportunities. These modules are connected through a route-weighted aggregation layer that yields final product-level shipping cost.

\subsection{Problem Formulation}
Under our simplified fulfillment setting, we set Boston as the single in-stock warehouse and all the order shipment originates from Boston and deliver through a single parcel carrier (FedEx).  We define routes at the destination-zone level. In practice, these destination zones are obtained from the FedEx zone locator: after specifying Boston ZIP 02108 as the shipping origin, each destination ZIP code is assigned a zone index ranging from 1 to 8. This carrier-defined zone index is used as the route dimension in our formulation. Accordingly, let pdenote a product, z$\in${1,...,8}a destination zone, and ta forecast period. The final product-level shipping cost estimate is given by

\begin{equation}
\hat{C}(p,t)=\sum_{z=1}^{8} w(z\mid p,t)\,\hat{c}(p,z,t)
\end{equation}

where w(z | t) is the forecasted route weight for zone z at time t and $\hat{c}$(p, z, t) is the estimated route-level shipping cost. Route-level cost is decomposed as

\begin{equation}
\hat{c}(p,z,t)=c^{(1)}(p,z,t)+\Delta c^{(2)}(p,z,t)-s^{\text{box}}(p,z,t)
\end{equation}

This route definition follows directly from our single-origin, single-carrier assumption, under which the effective routing choice is reduced to destination zone only. More generally, if the fulfillment system includes multiple warehouses and multiple carriers, the route space expands beyond zones alone to a cross-product of destination regions, warehouse choices, and carrier choices. In that setting, the number of possible routes for each product becomes $\lvert R \rvert = \lvert Z \rvert \times \lvert W \rvert \times \lvert K \rvert$, where $Z$ is the destination-region set, $W$ is the warehouse set, and $K$ is the carrier set. In the present paper, we intentionally restrict the environment to $\lvert W \rvert = 1$ and $\lvert K \rvert = 1$ in order to isolate the cost-estimation problem itself.

\subsection{Demand Forecasting and Multi-Stage Cost Estimation}
\begin{figure*}[!t]
\centering
\includegraphics[width=\textwidth]{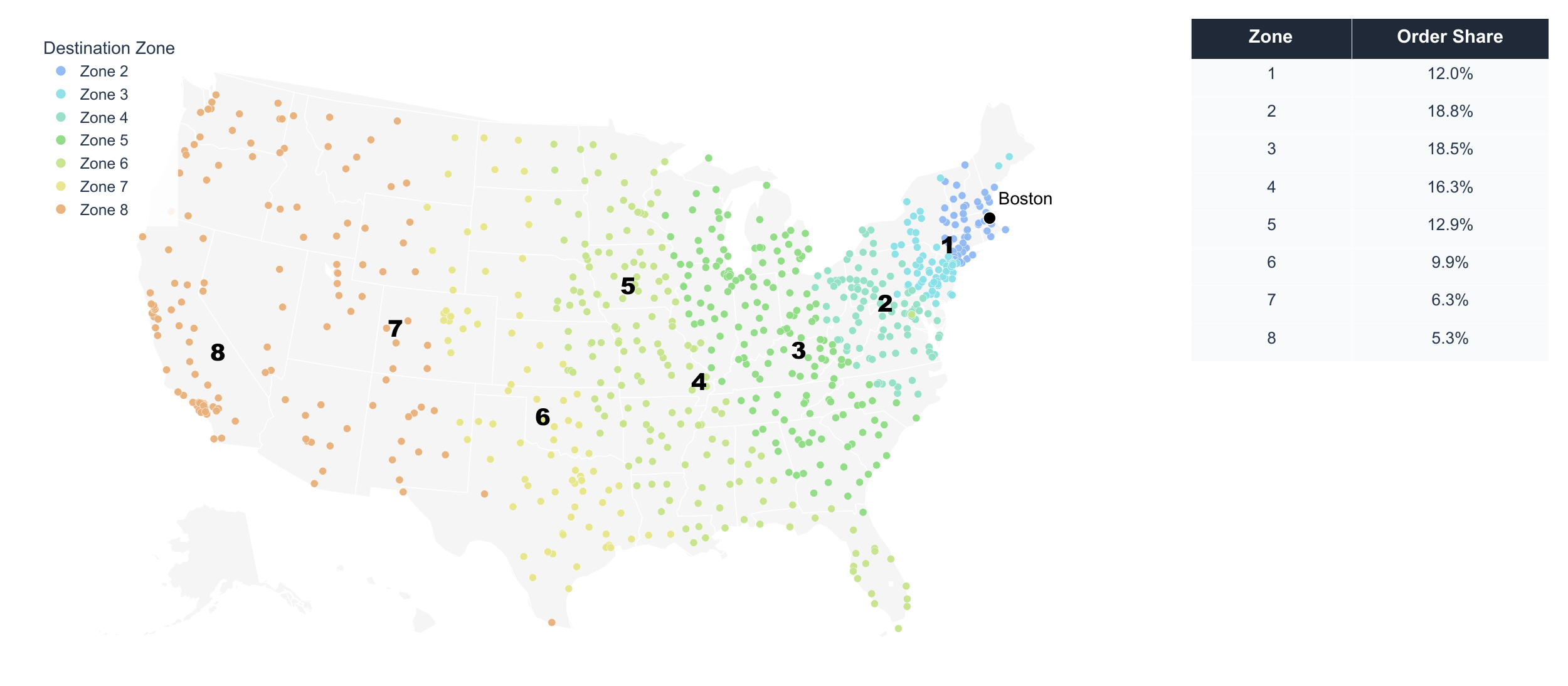}
\caption{Boston-origin destination zone map. Each point represents a sampled destination ZIP area colored by FedEx zone relative to the Boston origin. Larger point markers are used to improve geographic readability and visually highlight the spatial zone structure.}
\label{fig:zone-map}
\end{figure*}

The demand module estimates zone-level route weights from historical order flow. Rather than assuming a fixed destination distribution, we construct monthly zone shares and smooth them using short rolling windows. To assign each order to a route, we use the FedEx zone locator with Boston ZIP 02108 as the shipping origin and retrieve the carrier-defined zone label for each destination ZIP code. As shown in Figure~\ref{fig:zone-map}, this procedure produces a Boston-origin destination-zone map in which each destination ZIP is associated with one of the eight zone indices used in the formulation above. Orders are then mapped to destination zones through their destination ZIP codes, which enables both zone-level demand estimation and route weighting. Because destination patterns vary across product groups, we also estimate category-specific monthly zone shares and apply a hierarchical fallback strategy during inference.

\begin{figure}[!t]
\centering
\includegraphics[width=\columnwidth]{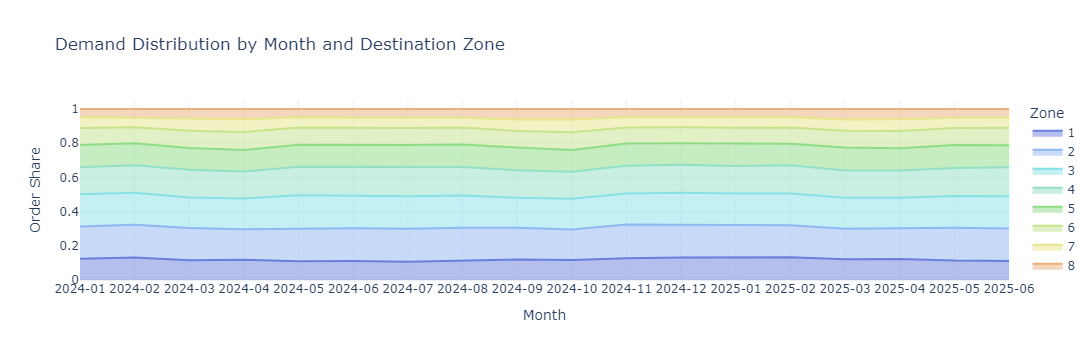}
\caption{Demand distribution by month and destination zone}
\label{fig:demand-distribution}
\end{figure}

After assigning orders to zones, we aggregate them by month to obtain the temporal demand distributions shown in Figure~\ref{fig:demand-distribution}; these distributions provide the route weights used by the downstream cost modules.

Stage 1 is designed as a structured baseline rather than a simple arithmetic lookup. Carrier rate-card information remains an important signal, but the final baseline estimate is learned through a regularized Ridge regression model that integrates billable weight, physical weight, dimensional weight, package dimensions, destination zone, synthetic rate-card lookup values, and surcharge-related flags. \cite{ref9} This stage is intended to capture the relatively stable pricing structure of the problem, including the parts of shipping cost that are directly tied to physical shipment characteristics and carrier pricing logic. Stage 2 then learns a nonlinear residual correction over the Stage 1 output using gradient boosting. By construction, this stage does not need to relearn the full pricing function, but instead focuses on the remaining nonlinear error not explained by the structured baseline. Finally, box consolidation is inferred from weak operational proxies such as same-day same-ZIP density, same-day same-zone density, family-level average quantity, size compatibility, package split risk, and a composite consolidation opportunity score. This final module targets latent operational savings that are difficult to encode through explicit pricing variables alone and are not directly observable at prediction time.

For each product, we construct an explicit route table containing all candidate destination-zone routes. Each route entry stores the route identifier, forecasted route weight, predicted route-level shipping cost, and weighted contribution to final expected cost. Because the final estimate is computed as the sum of route-level weighted contributions, the model can be audited at both the prediction level and the route-composition level. Taken together, the multi-stage design reflects a functional separation of sources of variation: Stage 1 models stable pricing structure, Stage 2 captures residual nonlinear error, and the consolidation module accounts for latent fulfillment effects. This separation reduces the need for a single model to jointly absorb pricing rules, demand patterns, and hidden operational effects in one end-to-end mapping. Viewed more generally, this formulation is not merely a weighted sum, but a structured decomposition of prediction uncertainty over latent routes. Before an order is placed, the final route realization is unknown, so product-level expected shipping cost is expressed through destination-zone demand shares and route-level pricing estimates rather than predicted directly from a single mixed feature space. Under the present single-origin, single-carrier setting, this decomposition is defined over destination-zone routes only, which keeps the formulation aligned with the simplified fulfillment environment studied in this paper.

\section{Dataset and Experimental Setup}
We evaluate the framework on an operationally grounded dataset constructed to emulate small-parcel shipping behavior under a simplified e-commerce setting. The dataset contains 250,000 order records, 260 products, and 18 months of transaction history. Each order record includes an order date, destination ZIP code, destination zone, product identifier, and purchased quantity. Product records contain physical attributes such as unit weight, length, width, and height, together with category and product-family identifiers. Table~\ref{tab:dataset} summarizes the dataset and experimental setup specifications.

The fulfillment setting is intentionally simplified. All shipments are assumed to originate from a single Boston warehouse, use a single parcel carrier, and ship under a standard service level. The synthetic pricing layer is built from a simplified FedEx-like rate-card structure indexed by destination zone and billable weight. We use a time-based split in which earlier months are used for training and the final three months are reserved for holdout evaluation. Importantly, this chronological split is used for backtesting rather than final deployment; once evaluation is complete, the production-facing version of the model can be refit on the full available history so that the most recent demand and surcharge patterns are retained. We report MAE, MAPE, and aggregate monthly error.

\begin{table}[htbp]
\caption{Dataset and setup summary}
\label{tab:dataset}
\centering
\begin{tabular}{@{}p{0.27\columnwidth}p{0.63\columnwidth}@{}}
\hline
Component & Specification \\
\hline
Orders & 250,000 line-item orders \\
Products & 260 products across 36 categories \\
Time span & 2024-01 to 2025-06 (18 months) \\
Fulfillment setting & Single Boston warehouse; single parcel carrier \\
Route granularity & 8 destination zones \\
Pricing layer & Zone x billable-weight rate card with surcharges \\
Consolidation signal & Latent savings inferred from weak operational proxies \\
Evaluation & Time-based holdout; MAE, MAPE, aggregate error \\
\hline
\end{tabular}
\end{table}

\section{Results}
\subsection{Main Performance}
\begin{figure}[!t]
\centering
\includegraphics[width=\columnwidth]{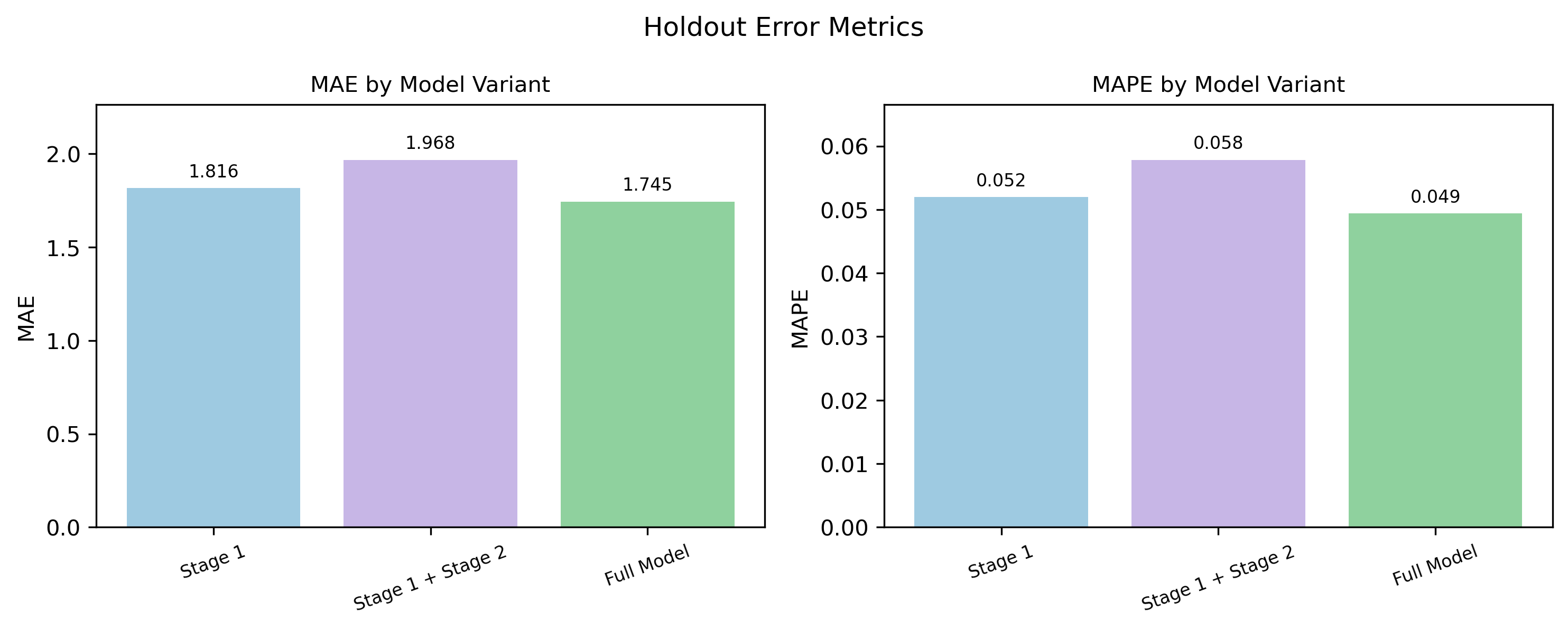}
\caption{Holdout MAE and MAPE across model variants}
\label{fig:holdout-metrics}
\end{figure}

We compare three variants of the pipeline: Stage 1 only, Stage 1 + Stage 2, and the full framework of stage 1 + stage 2 with proxy-based box-consolidation inference. The fee-card-informed Stage 1 model already provides a meaningful baseline because it captures structured pricing signals associated with billable weight, dimensions, surcharge triggers, and destination zone. Adding Stage 2 improves the estimate by correcting nonlinear residual error, while the full framework further benefits from the inferred consolidation effect. Overall, as reported in Table~\ref{tab:performance} and visualized in Figure~\ref{fig:holdout-metrics}, the full model achieves the best holdout performance, indicating that each additional stage contributes useful information beyond the structured baseline.

\begin{table}[htbp]
\caption{Holdout performance comparison across model variants}
\label{tab:performance}
\centering
\begin{tabular}{lcc}
\hline
Model Variant & MAE & MAPE \\
\hline
Stage 1 & 1.735 & 0.070 \\
Stage 1 + Stage 2 & 1.893 & 0.078 \\
Full Model & 1.649 & 0.064 \\
\hline
\end{tabular}
\end{table}

Figure 5 further illustrates this decomposition at the average holdout-order level, showing how the final estimate is formed progressively from the Stage 1 baseline through residual adjustment and consolidation savings.

\begin{figure}[!t]
\centering
\includegraphics[width=\columnwidth]{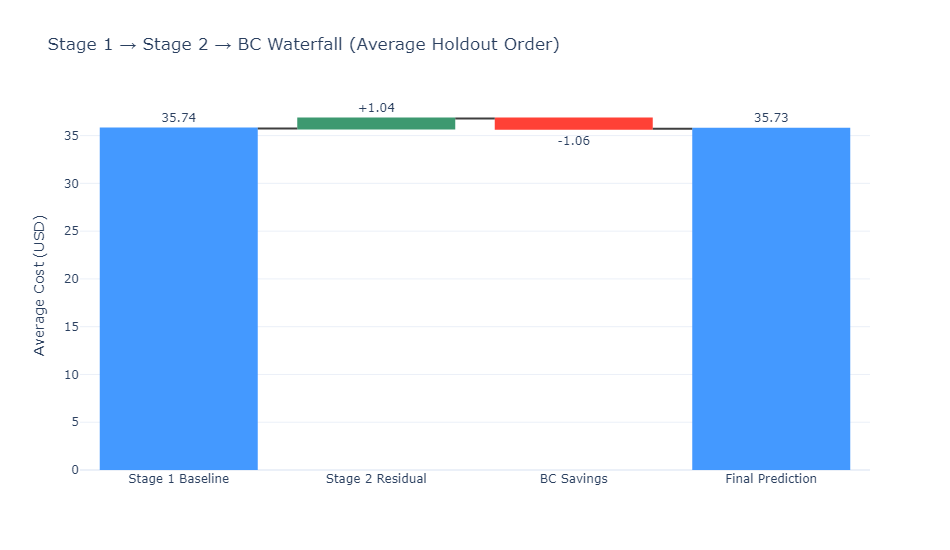}
\caption{Waterfall decomposition of the average holdout-order prediction, showing how the final estimate is obtained from the Stage 1 baseline through Stage 2 residual correction and box-consolidation savings.}
\label{fig:waterfall}
\end{figure}

\subsection{Temporal Error Analysis}
To examine calibration stability over time, we compute aggregate prediction error by month over the full 18-month horizon, defined as the ratio of total predicted shipping cost to total actual shipping cost minus one. This metric is especially useful because a model with acceptable per-order accuracy can still generate biased monthly totals, which would be problematic for pricing, budgeting, and gross-margin planning. As shown in Figure 6, the full model remains close to zero aggregate bias in most months, with fluctuations contained within a narrow range of roughly -0.14\% to +0.06\%. This full-history view provides a clearer picture of temporal calibration than a short holdout-only summary and suggests that the framework is not only accurate at the order level but also reasonably stable at the monthly aggregate level.

\begin{figure}[!t]
\centering
\includegraphics[width=\columnwidth]{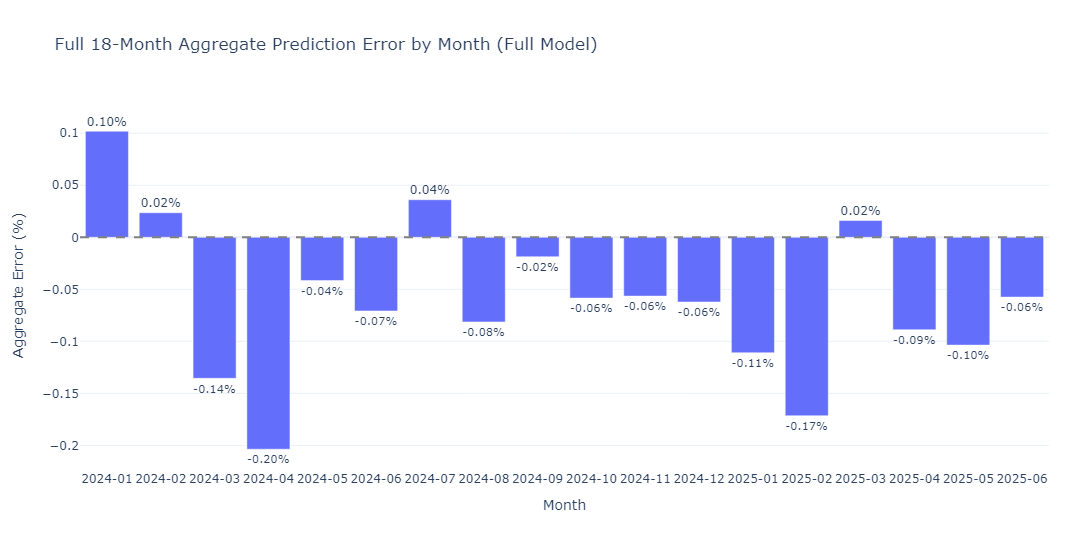}
\caption{Aggregate prediction error by month over the full 18-month horizon for the full model.}
\label{fig:aggregate-error}
\end{figure}

\subsection{Route-Level Interpretability}
One of the main advantages of the proposed framework is that it produces interpretable intermediate outputs instead of only final point predictions. The demand module yields monthly zone shares, while the route layer produces route weights, route costs, and weighted contributions. In practical terms, route cost and route contribution are not identical: some far-away zones contribute strongly because route costs are high even when demand share is modest, while some nearby zones dominate mainly because demand is concentrated there. This route-weighted interpretation is especially valuable for downstream pricing discussions because it ties expected cost back to both operational demand patterns and pricing mechanics.

\section{Discussion and Limitations}
The results support three main observations. First, pre-order shipping cost estimation benefits from being formulated as a route-based expectation problem rather than a single end-to-end regression task. The route-weighted view naturally connects destination demand and route-level pricing, improving interpretability and producing useful intermediate outputs such as weighted contribution breakdowns. Second, the fee-card-informed baseline is more reliable than either a pure lookup rule or an unconstrained black-box regressor. A further advantage of this decomposition is robustness under distribution shift. In practice, different parts of the fulfillment environment may change for different reasons and on different timescales. Demand-side patterns can shift because of seasonality or regional demand reallocation, while pricing-side conditions can change because of holiday peak surcharges, carrier fee updates, or rate-card revisions. In such settings, a modular design supports targeted refresh of the affected components: demand models can be updated when destination mix changes, while pricing logic or rate-card inputs can be revised when carrier pricing rules change. This reduces unnecessary relearning in components whose underlying mechanism remains stable. The same structure also makes business rule embedding more practical and auditable, since fee cards, dimensional pricing logic, billable-weight effects, and surcharge triggers can be injected explicitly into Stage 1 and updated locally when pricing rules or operational policies change.

At the same time, several limitations should be acknowledged. The evaluation uses an operationally grounded synthetic dataset and a simplified fulfillment network with a single origin, single carrier, and single service level, which reduces route complexity. In addition, the consolidation target is inferred through pseudo-label construction rather than learned from true shipment-level consolidation outcomes. A key next step is extending the route representation beyond a single Boston-origin zone map. In a multi-warehouse setting, maintaining a separate zone map for each origin would not scale well. A more scalable alternative is to introduce a shared regional partition of the U.S., where each region groups destinations that have similar effective fulfillment cost from the most likely serving origin. Under this view, region boundaries emerge near cost-indifference frontiers, where multiple fulfillment centers provide comparable shipping cost. The model could then learn, for each warehouse, its fulfillment mix over the same common set of regions and expand the route definition from product $\times$ zone to warehouse $\times$ region or warehouse $\times$ region $\times$ carrier. Such a representation would also align more naturally with upstream supplier-selection and fulfillment-allocation models that operate on a unified regional demand view. Beyond representational scalability, a multi-stage design could also support operational sequencing of model refreshes in richer fulfillment settings. For example, when regional demand distribution shifts over time, updated demand estimates could first inform warehouse-selection or fulfillment-allocation components, after which downstream route-level pricing modules could be refreshed if needed. Although this sequential update logic is not implemented in the present single-origin setting, it suggests how the same modular framework could extend to more realistic multi-warehouse or multi-carrier systems. Future extensions may also benefit from retrieval-augmented and LLM-assisted tooling. Dynamic-aware RAG systems suggest a way to combine static documentation with time-sensitive external knowledge when carrier rules, surcharge policies, or service constraints change over time \cite{ref10}. Related reranking-based retrieval pipelines for long financial documents point to a practical path for extracting structured evidence from carrier invoices or other billing documents, which could improve supervision for realized shipping cost and surcharge reconciliation \cite{ref11}. Finally, if future versions of the system incorporate short, noisy commerce text such as product notes or customer feedback, representation choice should be validated empirically rather than assuming transformer embeddings are always best, since classical embeddings can remain competitive on short industrial text \cite{ref12}. In addition, context-efficient frameworks for adapting smaller language models to domain-specific settings may provide a practical direction for lightweight operational document understanding and feature extraction when surcharge rules, billing descriptions, or carrier guidance are expressed in semi-structured text rather than clean tabular form \cite{ref13}. These studies are not directly comparable to the proposed framework, but they highlight future opportunities for extending shipping cost estimation systems with knowledge-aware components, such as policy retrieval, exception handling, and human-interpretable decision assistance.

\section{Conclusion}
This paper introduced RouteCost, a production-inspired multi-stage framework for pre-order shipping cost estimation in e-commerce. Instead of treating the task as a single regression problem, we formulated it as a route-weighted expectation problem and decomposed it into time-aware demand forecasting, fee-card-informed baseline pricing, residual correction, and proxy-based box-consolidation inference. The resulting framework offers a practical middle ground between rigid rule systems and monolithic black-box predictors. More broadly, the study suggests that operational cost-estimation problems benefit from structured decomposition: when pricing logic, temporal demand shift, and latent fulfillment effects coexist, a modular architecture can provide a more robust and explainable alternative to simpler approaches.

\bibliographystyle{ACM-Reference-Format}
\bibliography{references}

\end{document}